# Global Image Segmentation Process using Machine Learning algorithm & Convolution Neural Network method for Self-Driving Vehicles

Tirumalapudi Raviteja[1], Rajay Vedaraj .I.S[2]

[1]Research Scholar, School of Mechanical Engineering, Vellore Institute of Technology, Vellore, India

[2]Professor, School of Mechanical Engineering, Vellore Institute of Technology, Vellore, India

* Corresponding author: rajay@vit.ac.in

**Abstract:** In autonomous Vehicles technology Image segmentation was a major problem in visual perception. This image segmentation process is mainly used in medical applications. Here we adopted an image segmentation process to visual perception tasks for predicting the agents on the surrounding environment, identifying the road boundaries and tracking the line markings. Main objective of the paper is to divide the input images using the image segmentation process and Convolution Neural Network method for efficient results of visual perception. For Sampling assume a local city data-set samples and validation process done in Jupyter Notebook using Python language. We proposed this image segmentation method planning to standard and further the development of state-of-the art methods for visual inspection system understanding. The experimental results achieves 73% mean IOU. Our method also achieves 90 FPS inference speed and using a NVDIA GeForce GTX 1050 GPU.

1. **Introduction**: Object detection and Visual Perception technologies are important for Autonomous vehicles (AVs)[1]–[3]. Current scenario of AV's are using 6g technology locate accurate position and try to reduce latency position (< 1ms)[4]. This image segmentation process has a different applications, here we used in visual perception for AVs[5]–[8]. Early this segmentation process using in medical applications, later on that adapted to different areas. In particular some methods are developed for visual perception. For getting accurate results some sensors equipped in the autonomous vehicles such Lidar, Radar, Camera, IMU, etc. In global systems at all scenarios like at traffic places, high-way lane changes, urban roads and different weather conditions AV's take their own decision using visual perception systems. So that visual perception system is very significant process in AVs.

For many years, image segmentation was an intricate errand in Self-driving vehicles and it is also called semantic segmentation. Presently it is simpler with profound learning. Image segmentation is unique in relation to image order. In image order, it will just group protests that it has explicit marks for, for example, street, vehicle, and bike and so forth though image segmentation calculations will likewise portion obscure objects. Semantic segmentation understands images at each pixel level. All in all, Image segmentation is the errand of separating the picture into sections having a place with similar objects. Image characterization is the assignment of arranging what shows up in a picture into one out of a lot of predefined classes. Self-driving vehicles require a profound comprehension of their environmental factors. To help this, sensor outlines are utilized to perceive all agents in the road at a pixel-level precision. Utilizing image segmentation measures serves to characterize perceived items in images.

Machine learning is broadly used to discover the answers for different difficulties emerging in assembling self-driving vehicles[6], [9]. With the use of sensor information to control a vehicle, it is fundamental to improve the use of AI to achieve new undertakings[10]–[12]. The potential applications incorporate assessment of driver condition or driving situation characterization through information combination from various outside and inward sensors – like Lidar, radars, and cameras. In the self-driving vehicle, one of the significant undertakings of a K-means Clustering algorithm [13], [14]is consistent delivering of general climate and determining the progressions that are conceivable to these environmental factors to location of an Object, the Identification of an Object or acknowledgment object order and the Object Localization and Prediction of Movement of the agents as shows in Fig.1.

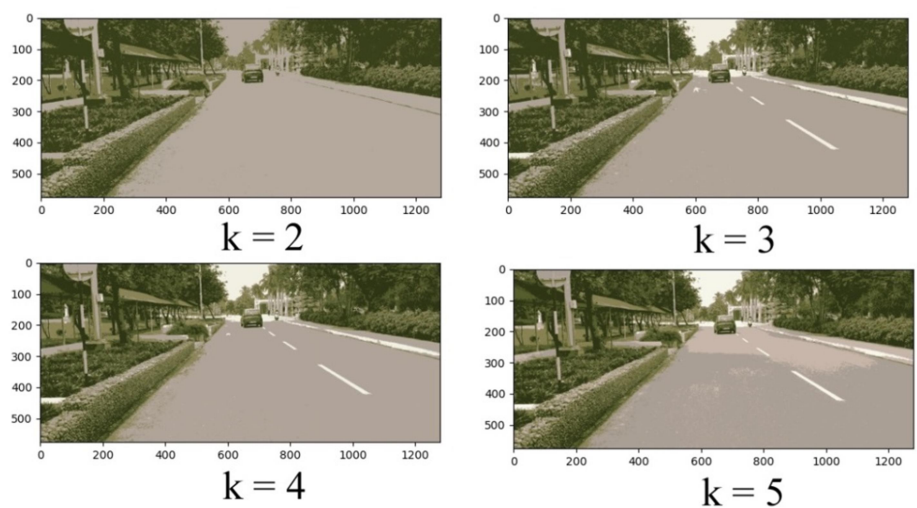

Fig 1. Sample image is classifying the 2, 3, 4 and 5 K-means Clusters

Deep Learning is one of the approaches to make self-driving is possible. Artificial Neural Networks (ANN) are data handling frameworks made out of straightforward preparing units, interconnected and acting in equal[12], [15]. The motivation is from the organic sensory systems. Sensor perception is one of the territories of Artificial Intelligence (AI) where they need to remove data out of pictures. Furthermore, with the tremendous requests and desires for individuals toward this field, it is ascending past convictions: beginning from object identification, design acknowledgment, activity acknowledgment, programmed direction, etc. Some papers have been distributed about it, particularly in Deep Learning and Convolutional Neural Network (Deep CNN). The CNN strategy has to a great extent in different fields: Image segmentation, AI, Computer vision and substantially more. In the Self-driving vehicle technology, the CNN technique has been for the most part utilized for object location and computerization dependent on the pictures as shows in Fig 2. Those CNN strategies are essentially utilized for object location and acknowledgment[6], [16]–[18].

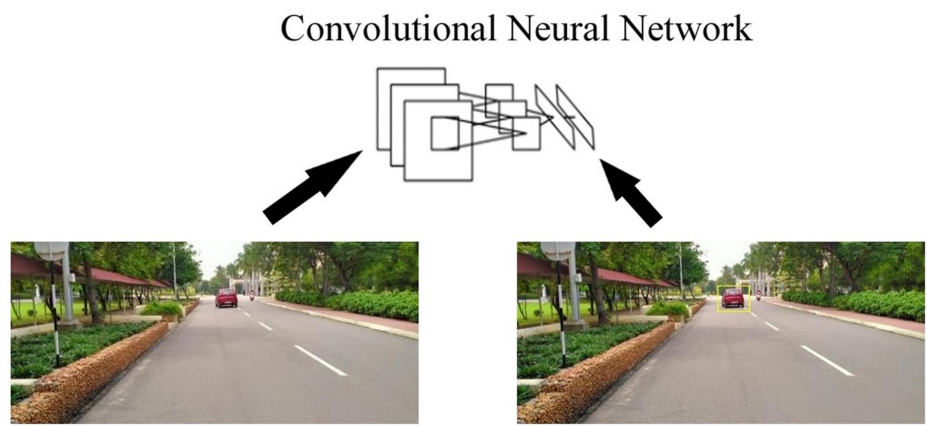

Fig 2. Image features are extracted and identified by Deep CNN. Car Image is identified and marked as a yellow colour box.

2. **Related to work:**

The self-driving car has several sub systems of Perception and decision making systems. These sub systems are detailed as shown in Fig 3.

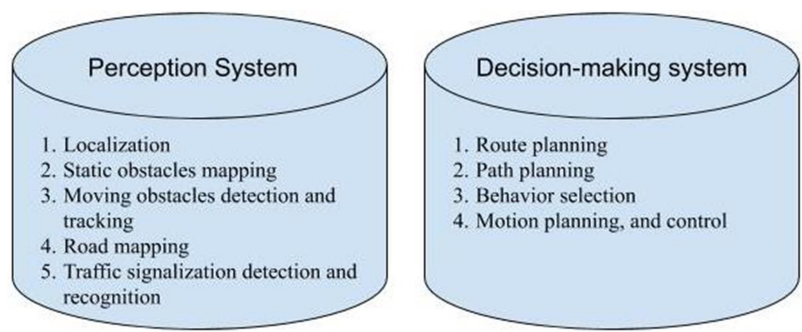

Fig 3. Sub systems of Perception systems and Decision-Making systems

Huval et al. [8] propose a neural-based strategy for recognition of moving agents utilizing the Overfeat CNN. Mutz et al. [8] labels moving agents and ego vehicles monitoring for a closely related application known as 'follow the ego vehicle. Jensen et al. [8] referred to vehicle detecting and recognition of traffic lights (Red, Green and Yellow). Sermanenu et al. [8] proposed Overfeat CNN work at predicting the distance from current state of the vehicle to ego vehicles. Caludine Badue et al. [8] presented a detailed survey about self-driving cars perception and decision making systems.

### 2.1. Image Segmentation:

Novozamsky et al. [19] propose a real time automated object labelling for CNN based image segmentation. Liand - chien chen et al. [9] address a convolution with unsampled filters for control the resolution of image at deep CNN and spatial pyramid polling (SPP) for divide the objects at multiple scales and combination of DCNNs and probabilistic graphical models for localization of object boundaries with 79.7% of Imou results. Di Feng et al. [20] address challenges, datasets and existing methods of the deep multi model object detection and semantic segmentation for autonomous driving. Sagar A et al. [21] represents a new neural network model of multi scale feature fusion for efficient semantic image segmentation and use for self-driving cars with 74.12% IOU results .

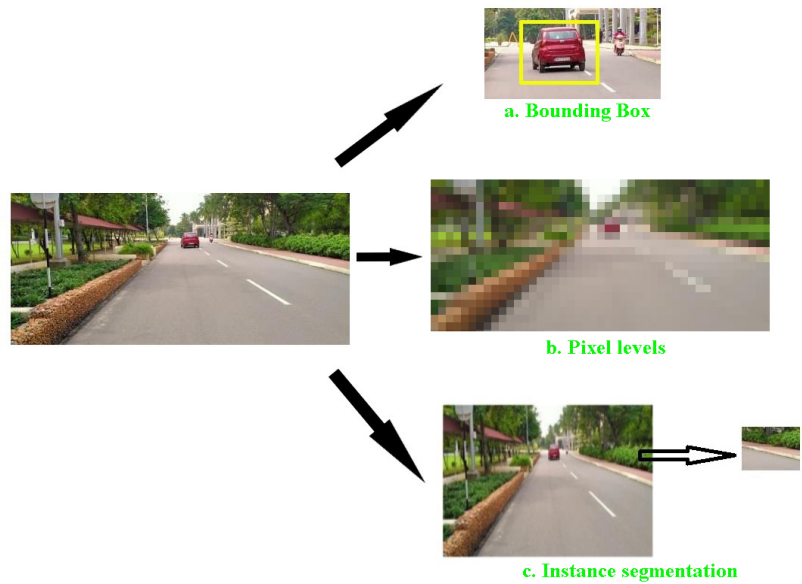

Fig 4. Image segmentation Process a) Bounding Box, b) Pixel level and c) Instance segmentation

### 2.2. Convolution Neural Network:

Quan Zhou et al. [22] present an attention-guided lightweight network (AGLNet) to employ the encoder and decoder architecture for real-time semantic segmentation. Mengyu Liu et al.

[6] proposed a lightweight feature pyramid encoding network (FPENet) for efficient accuracy and speed. Shervin et al. [5] address a systematic survey of image segmentation using deep learning in wide areas. Some of the well-known CNN architectures are AlexNet, DenseNet, MobileNet and ResNet[5], [23]. Michael Treml et al. [24] propose new deep network architecture for image segmentation with consists of active functions for high accuracy rate.

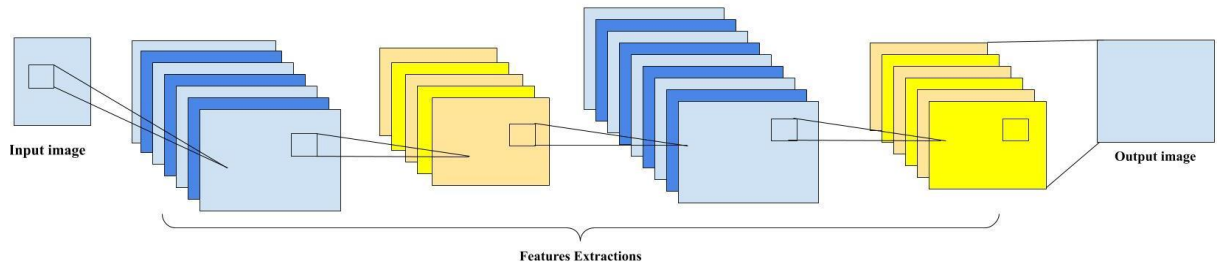

Fig 5. Simple CNN architecture

We summarize above related work and our Frame work contribution as follows:

- We proposed a new model for using the image segmentation process and Convolution Neural Network method for efficient results of visual perception.
- We present the K-means clustering layer details, optimization of our model.
- On evaluating our network model and datasets are the efficient mean and accuracy of MOU results > 100FPS.

3. **Our Frame Work:**

Our framework model consists of seven layers, as represented in Fig.6.

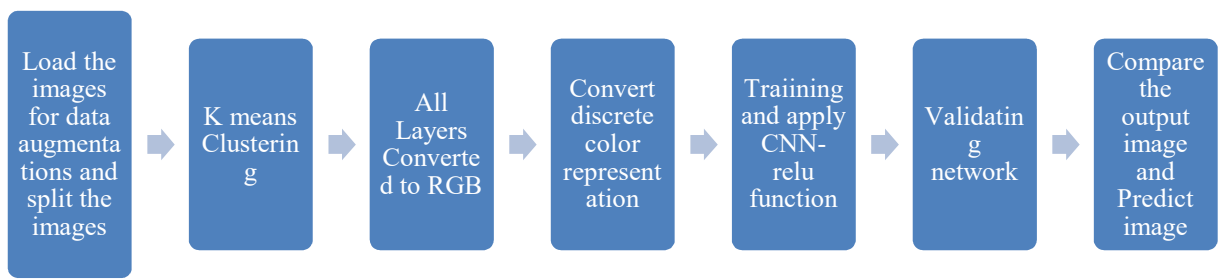

Fig 6. Phases of Our network model

The process starts with the install required package in python such as numpy, os, cv2, k means, random, Conv2d, leakyReLU, Adam and SGD [25]. Then load the datasets for data augmentation like flipping and rotating & split the images. After the data augmentation is divided as clustering by colors using K Means clustering. After the clustering process image layers to be converted as RGB images, it means representing the colors. After the colour representation process is classified as classes in image and generates the data. The generating data train by CNN. Then CNN Models are trained with 1000 epochs and validate the model.

The colour segmentation image and true images classes will be shown in Fig. 7

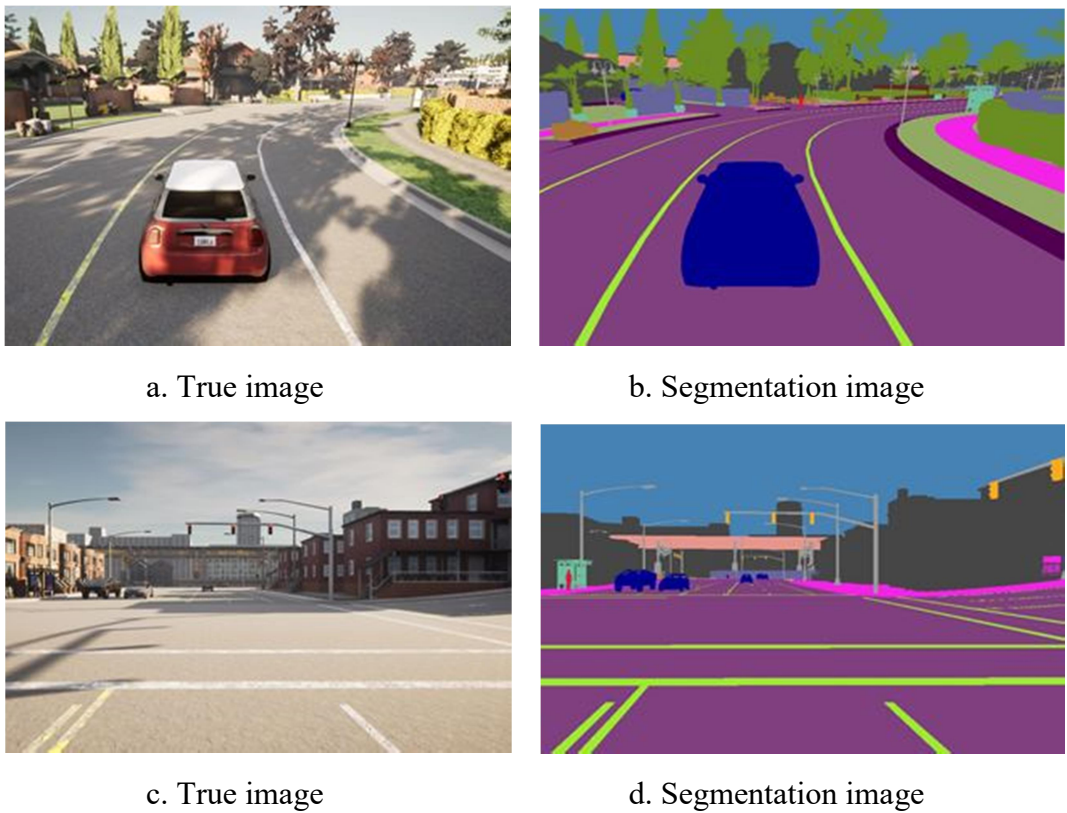

a. True image          b. Segmentation image

c. True image          d. Segmentation image

Fig 7. True image and Colour Segmentation image

Our CNN was built in Keras, the high-level API in TensorFlow, and it was trained on NVIDIA GEFORCE GTX 1050 graphics card. Training data and results are discussed in the next session.

## 4. Experiment Results:

Class IOU is calculated by average of pixels True positive (TP), False positive (FP) and False negative (FN) as shown in Table 1.

Intersection over union (IOU) = $\frac{Ture\ Positive}{True\ Positiv\ \ Positive+Fal\ \ Negitive}$

F(x , θ) = [Class_01, Class_02, -----, Class_12]

Table 1: Classification Of Segmentation image class Metrics

|  | Class _01 | Class _02 | Class _03 | Class _04 | Class _05 | Class _06 | Class _07 | Class _08 | Class _09 | Class _10 | Class _11 | Class _12 |
|---|---|---|---|---|---|---|---|---|---|---|---|---|
| TP | 4 | 3 | 3 | 3 | 4 | 3 | 3 | 4 | 4 | 3 | 4 | 3 |
| FP | 0 | 2 | 2 | 2 | 2 | 0 | 2 | 2 | 0 | 0 | 0 | 0 |
| FN | 2 | 0 | 2 | 0 | 0 | 2 | 0 | 0 | 2 | 2 | 2 | 2 |
| IOU | 0.6 | 0.6 | 0.4 | 0.6 | 0.6 | 0.6 | 0.6 | 0.6 | 0.6 | 0.6 | 0.6 | 0.6 |

The average mean values of all class values are as shown as in Fig 8.

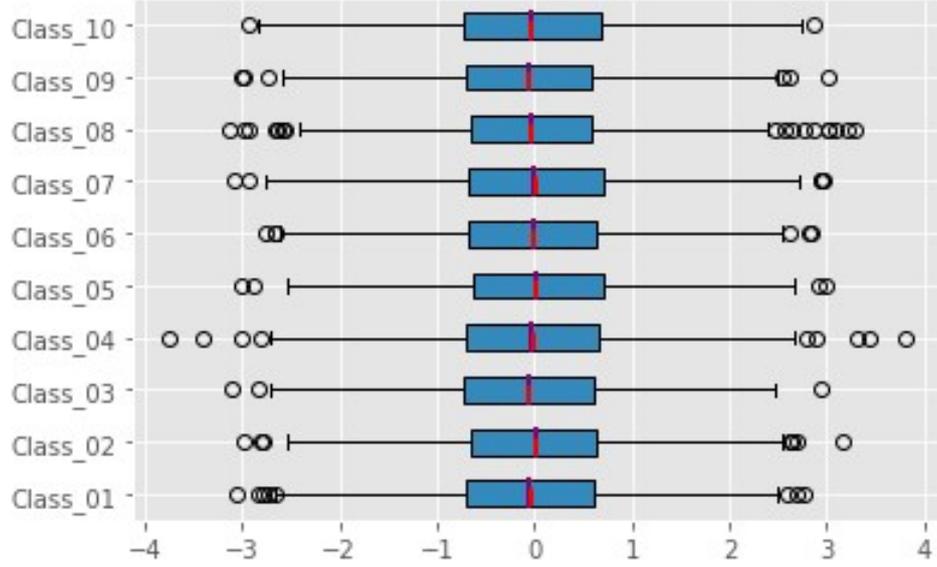

Fig 8. Mean of classes

All class values are Compare with the other networks are CGNet, ENet, ERFNet, FSCNet, FSCNN and DABNet. Except class_04, the remaining all classes' value is greater than the existing network values as shown in Table 3. Param, FPS and MIoU values also compared with the other models that values are also as shown in Table 4.

Table 2: Accuracy results of individual classes and comparison with other networks

| Method | Class _01 | Class _02 | Class _03 | Class _04 | Class _05 | Class _06 | Class _07 | Class _08 | Class _09 | Class _10 | Class _11 | Class _12 |
|---|---|---|---|---|---|---|---|---|---|---|---|---|
| CGNet | 90.8 | 79.8 | 28.1 | **95.3** | 81.9 | 73.2 | 41.6 | 32.9 | 81.3 | 52.9 | 53.9 | 68.4 |
| ENet | 92.4 | 75.1 | 24.2 | 93.5 | 74.5 | 96.7 | 31.2 | 21.5 | 78 | 42.6 | 46.2 | 64.1 |
| ERFNet | 91.6 | 77.45 | 26.15 | 94.4 | 78.2 | 84.95 | 36.4 | 27.2 | 79.65 | 47.75 | 50.05 | 66.25 |
| FSCNN | 68.7 | 58.09 | 19.61 | 70.80 | 58.65 | 63.71 | 27.30 | 20.40 | 59.74 | 35.81 | 37.54 | 49.69 |
| DABNet | 69.4 | 72.61 | 24.52 | 88.50 | 73.31 | 79.64 | 34.13 | 25.50 | 74.67 | 44.77 | 46.92 | 62.11 |
| Our Model | **93.2** | **82.1** | **58.4** | 85.1 | **90.2** | **95.1** | **55.4** | **68.2** | **90.8** | **62.1** | **70.2** | **82.78** |

Table 3: Comparison with other networks

| Method | Param | FPS | MIoU |
|---|---|---|---|
| CGNet | 0.57 | 97.5 | 65.7 |
| ENet | **0.34** | 92.4 | 57.5 |
| ERFNet | 2.12 | 142.5 | 69.2 |
| FSCNN | 1.16 | **247.6** | 55.6 |
| DABNet | 0.72 | 138.2 | 66.8 |
| Our Model | 1.47 | 92.4 | **72.4** |

Optimization of our CNN model, assume that the image classification to as X classes. The output probability sample is that,

$$P_n = [P_1, P_2, \_\_\_, P_x]^T \qquad (1)$$

When the ground truth label index is $g_n$, the output probability as,

$$Where, P_n^* = [P_1^*, P_2^*, \_\_\_, P_x^*]^T \qquad (2)$$

$$p_n^* \{ 1 \; if \; x == g_n, 0 \; other \; wise \qquad (3)$$

## 5. Conclusion:

In this paper, we proposed a global image semantic segmentation network. Its attention to the feature of the image and results are validated by its performance. We present optimization details of our network model. The Mean IoU of our model is 72.4 and the segmentation process is also > 100FPS, it's compared to other network models and very much better than in accuracy rate. This CNN model is suitable for perception task in self-driving vehicles.